\documentclass{article}
\usepackage{spconf,amsmath,graphicx}
\usepackage{subfigure}
\usepackage{bm}
\usepackage{cite}

\title{CONTRASTIVE-CENTER LOSS FOR DEEP NEURAL NETWORKS} 
\name{Ce Qi$^1$, Fei Su$^{1,2}$\thanks{}
}

\address{$^1$School of Information and Communication Engineering\\
        $^2$Beijing Key Laboratory of Network System and Network Culture\\
        Beijing University of Posts and Telecommunications, Beijing, China\\
}

\begin{document}
%
\maketitle
\begin{abstract}
The deep convolutional neural network(CNN) has significantly raised the performance of image classification and face recognition. Softmax is usually used as supervision, but it only penalizes the classification loss. In this paper, we propose a novel auxiliary supervision signal called contrastive-center loss, which can further enhance the discriminative power of the features, for it learns a class center for each class. The proposed contrastive-center loss simultaneously considers intra-class compactness and inter-class separability, by penalizing the contrastive values between: (1)the distances of training samples to their corresponding class centers, and (2)the sum of the distances of training samples to their non-corresponding class centers. Experiments on different datasets demonstrate the effectiveness of contrastive-center loss.
\end{abstract}
\section{Introduction}
\label{sec:intro}
Recently, deep neural networks have achieved state of the art performance on different tasks such as visual object classification\cite{krizhevsky2012imagenet,simonyan2014very,szegedy2015going,he2015delving,he2015deep} and recognition\cite{taigman2014deepface,sun2014deep_0,sun2014deep_1,sun2015deeply,schroff2015facenet,taigman2015web,wen2016latent,wen2016discriminative}, showing the power of the discriminative features.

In general visual classification and recognition task, deep convolutional neural networks(CNN)\cite{krizhevsky2012imagenet,simonyan2014very,szegedy2015going,he2015deep} are usually chosen. For the discriminative features extracted from CNN, the performance is usually much higher than other traditional machine learning algorithms. Usually, the CNN maps images to high dimension space to let the softmax or SVM easy to classify the images to a certain class. The softmax loss only penalizes the classification loss, and does not consider the intra-class compactness and inter-class separability explicitly.

Recently, there are some works learning with even more discriminative features to further improve the performance of CNN. Some researchers use deeper, wider and more complex network structures to obtain better features, such as \cite{he2015deep}, in which the authors train a very deep neural network with some training tricks to make the network converge to get more discriminative features, but the training is relatively harder and not that effective. There are also some other efforts on new non-linear activations\cite{goodfellow2013maxout,he2015delving}, dropout\cite{krizhevsky2012imagenet} and batch normalization\cite{ioffe2015batch} to make the network perform better.

Another kind of strategy to obtain more discriminative features is to use auxiliary loss to train the neural network, such as contrastive loss\cite{sun2014deep_1}, triplet loss\cite{schroff2015facenet} and center loss\cite{wen2016discriminative}. The three new losses are proposed for the purpose of enforcing better intra-class compactness and inter-class separability. The contrastive loss and triplet loss do really improve the quality of features extracted from the network. The triplet needs carefully pre-selected triple samples consisted of two same people's face images and one different person's face image. And the selection of triple samples is significant for it will influence the result of training. The contrastive loss chooses couple sample pairs to get the loss, so contrastive loss needs careful pre-selection, too. What's more, if all possible training samples combinations are chosen, the number of training pairs and triplets would theoretically go up to O($N^2$), where N is the total number of training samples. The center loss\cite{wen2016discriminative}, which learns a center for each class and penalizes the distances between the deep features and their corresponding class centers, is a new novel loss to enforce extra intra-class compactness. However, the center loss does not consider the inter-class separability.

In this paper, we propose the contrastive-center loss, which learns a center for each class. This new loss will simultaneously consider intra-class compactness and inter-class separability by penalizing the contrastive values between: (1)the distances of training samples to their corresponding class centers, and (2)the sum of the distances of training samples to their non-corresponding class centers. The training process is simple because the contrastive-center loss does not need pre-selected sample pairs or triples.

Experiments and visualizations show the effectiveness of our proposed contrastive-center loss. The experiments on MNIST\cite{lecun1998mnist} and CIFAR10\cite{krizhevsky2009learning} demonstrate the effectiveness of contrastive-center loss on classification task. And the experiments of face recognition on LFW\cite{LFWTech} demonstrate the effectiveness of contrastive-center loss on recognition task.



\section{PROPOSED METHOD}
\label{sec:proposed method}

In this section, we introduce center loss and indicate its weakness of only considering intra-class compactness. Then the proposed contrastive-center loss is described, which simultaneously considers the intra-class compactness and inter-class separability. Using softmax loss assisted with our contrastive-center loss to train a deep neural network will do really boost the performance of the network.

\subsection{Center loss}
\label{ssec:center loss}

The features extracted from the deep neural network trained under the supervision of softmax loss are separable but not that discriminative enough, since they show significant intra-class variations, as shown in Fig. \ref{fig:visualization of softmax loss on mnist}. Based on the phenomenon, authors in \cite{wen2016discriminative} develop an effective loss function to improve the power of the deep features extracted from deep neural networks. Center loss minimizes the intra-class distances while keeping the features can be classified into right classes by softmax. Eq. (\ref{equ:center loss}) gives the center loss function.

\begin{equation}\label{equ:center loss}
L_c=\frac{1}{2}\sum_{i=1}^m \|x_i-c_{y_i}\|_2^2
\end{equation}
Where $L_c$ denotes the center loss. $m$ denotes the number of training samples in a min-batch. $x_i \in R_d$ denotes the $i$th training sample. $y_i$ denotes the label of $x_i$.  $c_{y_i} \in R_d$ denotes the $y_i$th class center of deep features. $d$ is the feature dimension.

When training the deep neural networks, authors in \cite{wen2016discriminative} adopt the joint supervision of softmax loss and center loss to train the networks, as formulated in Eq. (\ref{equ:softmax loss and center loss}).

\begin{equation}\label{equ:softmax loss and center loss}
L = L_s + \lambda L_c
\end{equation}
Where $L$ denotes the total loss of deep neural network. $L_s$ denotes the softmax loss. $L_c$ denotes the center loss. $\lambda$ denotes the scalar used for balancing the two loss functions.

\textbf{The weakness of center loss}: Discriminative features should have better intra-class compactness and inter-class separability. The center loss uses loss function Equation \ref{equ:center loss} to penalize big intra-class distances. However, center loss does not consider the inter-class separability. It will make the distances of different classes not that far, as show in Fig. \ref{fig:visualization of center loss on mnist}. As we know, if the distances of different classes is far enough, the features will more discriminative for the better inter-class separability. In addition, for the center loss just penalizes big intra-class distances, does not consider inter-class distances, the changing of inter-class is small, meaning the positions of class centers will be slightly changed through all the training process. As a result, if the network initializes the class centers using a relatively smaller variance, it will result in the smaller distances between class centers after training because the center loss function only penalize the big intra-class distances without considering the inter-class distances.

\subsection{Contrastive-center loss}
\label{ssec:contrastive-center loss}

As mentioned in section \ref{ssec:center loss}, the weakness of center loss is that it does not consider the inter-class separability. So, we propose a new loss function to consider the intra-class compactness and inter-class separability simultaneously by penalizing the contrastive values between: (1)the distances of training samples to their corresponding class centers, and (2)the sum of the distances of training samples to their non-corresponding class centers. Formally, it is defined as illustrated in Eq.\ref{equ:contrastive-center loss}.

\begin{equation}\label{equ:contrastive-center loss}
L_{ct-c}=\frac{1}{2}\sum_{i=1}^m\frac{ \|x_i-c_{y_i}\|_2^2}{(\sum_{j=1,j\neq{y_i}}^k\|x_i-c_j\|_2^2) +\delta}
\end{equation}
Where $L_{ct-c}$ denotes the contrastive-center loss. $m$ denotes the number of training samples in a min-batch. $x_i \in R_d$ denotes the $i$th training sample with dimension $d$. $d$ is the feature dimension. $y_i$ denotes the label of $x_i$. $c_{y_i} \in R_d$ denotes the $y_i$th class center of deep features with dimension $d$. $k$ denotes the number of class. $\delta$ is a constant used for preventing the denominator equal to $0$. In our experiments, we set $\delta=1$ by default.


Obviously, the contrastive-center loss can be used in deep neural network directly and the network will be trained as general deep neural network. The class centers $c_{y_i}$ will be updated through the training process. Comparing with those in the center loss, the class centers of our proposed contrastive-center loss will be updated to a more discrete distribution for the existence of penalization for too small distances between different class centers.

In this method, we update the class centers based on mini-batch, for it is not possible to update the centers based on the entire training set. And to make the training process is more stable, we use a scalar $\alpha$ to control the learning rate of class centers.

In each iteration, the deep neural network updates the class centers and network parameters simultaneously. The derivative of $L_{ct-c}$ with respect to $x_i$ and derivative of $L_{ct-c}$ with respect to $c_{n}$ is illustrated in Eq. (\ref{equ:derivative to x_i of contrastive-center loss}) and Eq. (\ref{equ:derivative to c_y_i of contrastive-center loss}) respectively. The two derivatives are used to update the parameters of deep neural networks and class centers respectively.

\begin{equation}\label{equ:derivative to x_i of contrastive-center loss}
\resizebox{0.9\hsize}{!}
{$
\begin{split}
\frac{\partial L_{ct-c}}{\partial x_i}= & \frac{x_i-c_{y_i}}{(\sum_{j=1,j\neq{y_i}}^k\|x_i-c_j\|_2^2) +\delta} \\
& -\frac{\|x_i-c_{y_i}\|_2^2 \sum_{j=1,j\neq{y_i}}^k(x_i-c_j)}{[(\sum_{j=1,j\neq{y_i}}^k\|x_i-c_j\|_2^2) +\delta]^2}
\end{split}
$}
\end{equation}

\begin{equation}\label{equ:derivative to c_y_i of contrastive-center loss}
\resizebox{0.9\hsize}{!}
{$
\frac{\partial L_{ct-c}}{\partial c_n}=\sum_{i=1}^m
  \begin{cases}
    \frac{c_{y_i-x_i}}{(\sum_{j=1,j\neq{y_i}}^k\|x_i-c_j\|_2^2) +\delta}    &\mbox{if $y_i = n$}\\
    \frac{(x_i-c_n)\|x_i-c_{y_i}\|_2^2}{[(\sum_{j=1,j\neq{y_i}}^k\|x_i-c_j\|_2^2) +\delta]^2}    &\mbox{if $y_i \neq n$}
  \end{cases}
$}
\end{equation}

In Eq. (\ref{equ:derivative to x_i of contrastive-center loss}) and Eq. (\ref{equ:derivative to c_y_i of contrastive-center loss}), the meaning of symbols are the same as those in Eq. (\ref{equ:contrastive-center loss}), except that $n=1,...,m$ denotes the current class center`s serial number.

\section{EXPERIMENTS}
\label{sec:experiments}

To verify the effectiveness of the contrastive-center loss, we evaluate the experiments on two typical visual tasks: visual classification and face recognition. The experiment results demonstrate our contrastive-center loss can not only improve the accuracy on classification, but also boost the performance on visual recognition. In visual classification, we use two wildly used dataset(MNIST\cite{lecun1998mnist} and CIFAR10\cite{krizhevsky2009learning}). In face recognition, the LFW\cite{LFWTech} dataset is used. We implement the contractive-center loss and do the experiments using the Caffe library\cite{jia2014caffe}.

\begin{figure}
\centering
\subfigure[softmax loss]
{
\label{fig:visualization of softmax loss on mnist}
\includegraphics[width=0.17\textwidth]{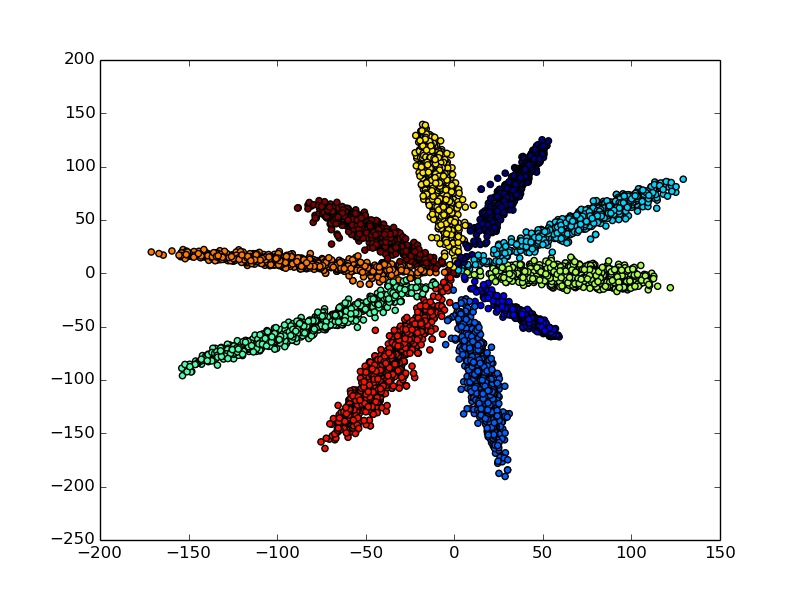}
}
\hspace{0in}
\subfigure[center loss]
{
\label{fig:visualization of center loss on mnist}
\includegraphics[width=0.17\textwidth]{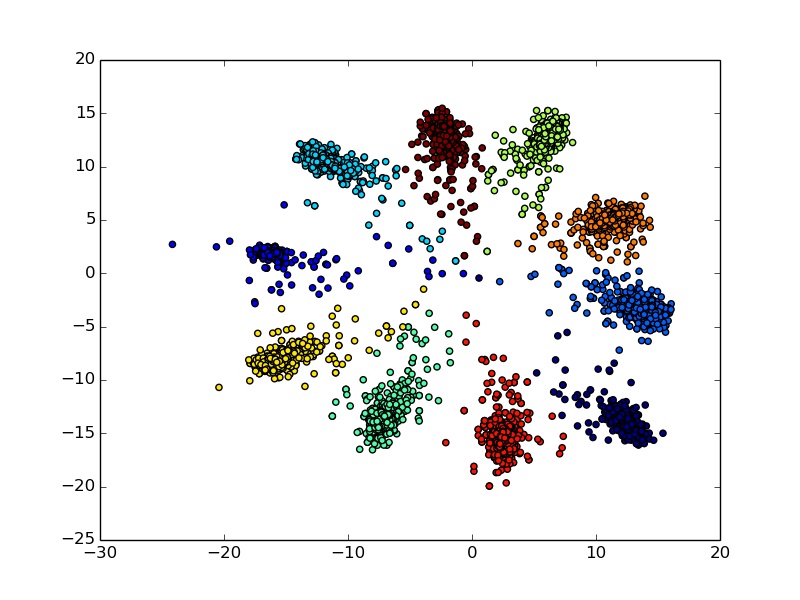}
}
\hspace{0in}
\subfigure[our contrastive-center loss]
{
\label{fig:visualization of contrastive-center loss on mnist}
\includegraphics[width=0.35\textwidth]{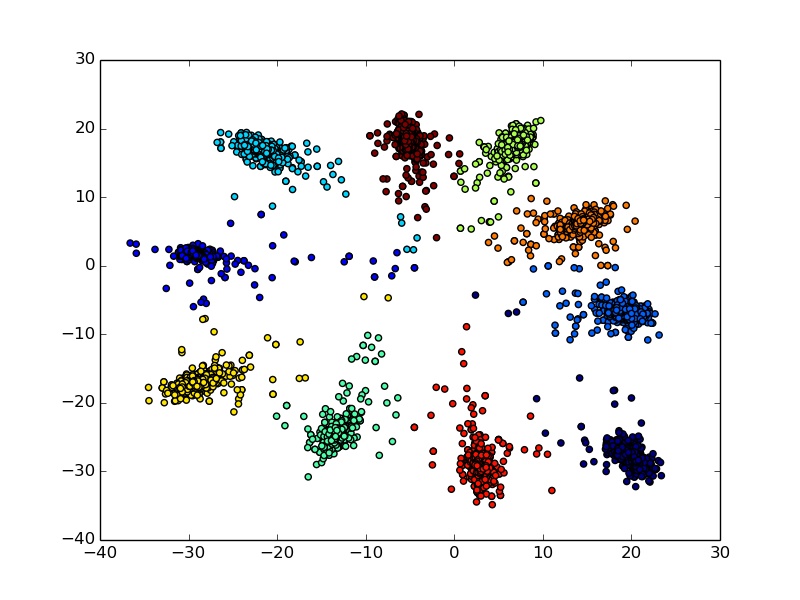}
}
\caption{Visualization of MNIST. Note: The domain of coordinates axis of the visualization of contrastive-center loss is bigger than center loss's. The average L2 distance of class centers to the center of class centers is about $10$ to $15$. The average L2 distance of class centers to the center of class centers is about $50$. The average L2 distance of class centers to the center of class centers is obvious, which indicates the contrastive-center loss's is about $3.3$ to $5$ times of the center loss's.}\label{fig:visualization of mnist}
\end{figure}


\subsection{Experiments on MNIST and visualiztion}
\label{ssec:experiments MNIST and visualiztion}

The MNIST\cite{lecun1998mnist} are consisted of $60,000$ training images and $10,000$ testing images in total. The images are all hand written digits $0-9$ in $10$ classes which are $28\times28$ in size.

The network used in this experiments are the same as the network used for MNIST in \cite{wen2016discriminative}. The network are modified from LeNets\cite{lecun1998gradient} to a deeper and wider network, but reduce the output number of the last hidden layer to 2, meaning the dimension of the deep features is 2, which is easy to be plot in 2-D surface for visualization. The details of the network architecture are given in Table \ref{table:lenet++}. Note that we set loss weight $\lambda=0.1$ for $L_{ct-c}$ there.

\begin{table}
\centering
\caption{The CNNs architecture we use for MNIST and visualization is same as \cite{wen2016discriminative}, called LeNets++. $(5,32)_{/1,2}\times2$ denotes $2$ cascaded convolution layers with $32$ filters of size $5\times5$, where the stride and padding are $1$ and $2$ respectively. $2_{/2,0}$ denotes the max-pooling layers with grid of $2\times2$, where the stride and padding are $2$ and $0$ respectively. In LeNets++, Parametric Rectified Linear Unit (PReLU)\cite{he2015delving} is used as the nonlinear unit.}\label{table:lenet++}
\scalebox{0.6}
{
\begin{tabular}{|*{8}{c|}}
\hline
\multicolumn{1}{|c|}{}&\multicolumn{2}{c|}{stage 1}&\multicolumn{2}{|c|}{stage 2}&\multicolumn{2}{|c|}{stage 3}&\multicolumn{1}{|c|}{stage 4}\\
\hline
\hline
Layer&conv&pool&conv&pool&conv&pool&FC\\
\hline
LeNets & $(5,20)_{/1,0}$ & $2_{/2,0}$ & $(5,50)_{/1,0}$ & $2_{/2,0}$ & {} & {} & 500 \\
\hline
LeNets++ & $(5,32)_{/1,2}\times2$ & $2_{/2,0}$ & $(5,64)_{/1,2}\times2$ & $2_{/2,0}$ & $(5,128)_{/1,2}\times2$ & $2_{/2,0}$ & 2 \\
\hline
\end{tabular}
}
\end{table}

When training and testing LeNet++, we only use original training images and original testing images without any data augmentation. The result is shown in Table \ref{table:result on mnist}. Contrastive-center loss boosts accuracy of $0.37\%$ compared to softmax loss and $0.23\%$ compared to center loss respectively.

\begin{table}
\centering
\caption{Classification accuracy (\%) on MNIST dataset.}\label{table:result on mnist}
\begin{tabular}{|*{2}{c|}}
\hline
Method&Accuracy(\%) \\
\hline
\hline
Softmax & $98.8$ \\
\hline
Center loss & $98.94$ \\
\hline
\textbf{Our contrastive-center loss} & $\bm{99.17}$ \\
\hline
\end{tabular}
\end{table}

We then visualize the deep features of the last hidden layer(the output number is $2$) of LeNet++. All the features are extracted using the $10,000$ testing images as input. The visualization is shown in Fig. \ref{fig:visualization of mnist}. We can observe that:

(1) Under the single supervision signal of softmax loss, the features are separable, but with significant intra-class variations.

(2) The center loss makes the deep features have better intra-class compactness. But the inter-class separability is not good enough. The average L2 distance of class centers to the center of class centers is about $10$ to $15$.

(3) The contrastive-center loss simultaneously achieves good intra-class compactness and inter-class separability. The average L2 distance of class centers to the center of class centers is about $50$.

(4) The contrastive-center loss's average L2 distance of class centers to the center of class centers is about $3.3$ to $5$ times of the center loss's, showing that the contrastive-center loss gets better inter-class separability than the center loss.

\subsection{Experiments on CIFAR10}
\label{ssec:experiments on CIFAR10}

The CIFAR10 dataset \cite{krizhevsky2009learning} is consisted of 10 classes of natural images with $50,000$ training images and $10,000$ testing images. Each image is RGB image of size $32\times32$.

We use 20-layer ResNet\cite{wen2016discriminative} in the experiments. Following the commonly used strategy, we do data augmentation in training, and in testing, there is no data augmentation. We follow the standard data augmentation in \cite{wen2016discriminative} for training: $4$ pixels are padded on each side, and a $32\times32$ crop is randomly sampled from the padded images or its horizontal flip. In testing, we only evaluate the single view of the original $32\times32$ testing images. Note that we set loss weight $\lambda=0.1$ for $L_{ct-c}$ there.

The result is shown in Table \ref{table:result on cifar10}. We can observe that:

(1) The center loss makes the net's accuracy increased by $0.4\%$ compared with the net's only supervised under softmax loss.

(2) Our contrastive-center loss makes the net's accuracy increased by $1.2\%$ compared with the net only supervised under softmax loss.

(3) Our contrastive-center loss gets better result than the center loss with accuracy gain of $0.35\%$ on CIFAR10.

\begin{table}[htb]
\centering
\caption{Classification accuracy (\%) on CIFAR10 dataset.}\label{table:result on cifar10}
\scalebox{0.7}
{
\begin{tabular}{|*{2}{c|}}
\hline
Method&Accuracy(\%) \\
\hline
\hline
20-layer ResNet\cite{he2015deep} & $91.25$ \\
\hline
20-layer ResNet(our implementation based on center loss\cite{wen2016discriminative}) & $92.1$ \\
\hline
\textbf{20-layer ResNet(our contrastive-center loss)} & $\bm{92.45}$ \\
\hline
\end{tabular}
}
\end{table}



\subsection{Experiments on LFW}
\label{ssec:experiments on LFW}

To further demonstrate the effectiveness of our contrastive-center loss, we conduct the experiments on LFW dataset\cite{LFWTech}. The dataset collects $13,233$ face images from $5749$ persons from uncontrolled conditions. Following the unrestricted with labeled outside data protocol\cite{LFWTech}, we
train on the publicly available CASIA-WebFace\cite{yi2014learning} outside dataset (490k labeled face images belonging to over $10,000$ individuals) and test on the $6,000$ face pairs on LFW. The training data is cleaned for wrong collected images. People overlapping between the outside training data and the LFW testing data are excluded. As preprocessing, we use MTCNN\cite{7553523} to detect the faces and align them based on $5$ points.

Then we train a single network for feature extraction. For good comparison, we use the network released by center loss\cite{wen2016discriminative}, which is called FRN(or FaceResNet) in later. Note that the network released by center loss is not the network they use in the paper\cite{wen2016discriminative}. Based on the network publicly available, we re-implement the training process of center loss and get better result than the models released by the author. We also train the networks under the supervision of softmax loss and our contrastive-center loss jointly. In feature extraction, like in \cite{wen2016discriminative}, the original image and its flip one are used to get two feature vectors and concatenate them as the final feature. Note that we set loss weight $\lambda=1$ for $L_{ct-c}$ there.

The result is shown in Table \ref{table:result on lfw}. We can observe that:

(1) We train the FRN(FaceResNet) only with small data(CASIA webface cleaned, $0.455,594M$). And the accuracy is comparable to the current state-of-art CNNs.

(2) FRN trained with our contrastive-center loss boosts the accuracy on LFW of $1.21\%$, $0.25\%$ and $0.13\%$ compared respectively with FRN trained with softmax loss only,  model released by center loss\cite{wen2016discriminative} and our re-implementation of center loss.

\begin{table}
\centering
\caption{Verification accuracy (\%) on LFW dataset. $*$ denotes the outside data is private (not publicly available).}
\scalebox{0.6}
{
\begin{tabular}{|*{4}{c|}}
\hline
Method&Images&Networks&Accuracy(\%) \\
\hline
\hline
DeepFace\cite{taigman2014deepface} & $4M$ & $3$ & $97.35$ \\
\hline
Fusion\cite{taigman2015web} & $10M$ & $5$ & $98.37$ \\
\hline
SeetaFace\cite{liu2016viplfacenet} & $0.5M$ & $1$ & $98.60$ \\
\hline
SeetaFace(Full)\cite{liu2016viplfacenet} & $0.5M$ & $1$ & $98.62$ \\
\hline
DeepID-2+\cite{sun2015deeply} & $-$ & $1$ & $98.70$ \\
\hline
DeepFR\cite{parkhi2015deep} & $2.6M$ & $1$ & $98.95$ \\
\hline
Yi et al., 2014\cite{yi2014learning} & $0.494,414M$ & $1$ & $97.73$ \\
\hline
Ding \& Tao, 2015\cite{ding2015robust} & $0.494,414M$ & $1$ & $98.43$ \\
\hline
\hline
FRN(trian with softmax loss only) & $0.455,594M$ & $1$ & $97.47$ \\
\hline
FRN(model released by center loss\cite{wen2016discriminative}) & $0.494,414M$ & $1$ & $98.43$ \\
\hline
FRN(retrain with center loss\cite{wen2016discriminative}) & $0.455,594M$ & $1$ & $98.55$ \\
\hline
\textbf{FRN(our contrastive-center loss)} & \bm{$0.455,594M$} & \bm{$1$} & \bm{$98.68$} \\
\hline
\end{tabular}
}
\label{table:result on lfw}
\end{table}

\section{CONCLUSION}
\label{sec:conclusion}

We proposed a contrastive-center loss for deep neural networks. The contrastive-center loss simultaneously considers intra-class compactness and inter-class separability, by penalizing the contrastive values between (1)the distances of training samples to their corresponding class centers, and (2)the sum of the distances of training samples to their non-corresponding class centers. More appealingly, the contrastive-center loss has very clear intuition and geometric interpretation. The experimental results on several benchmark datasets prove the effectiveness of the proposed contrastive-center loss.


\bibliographystyle{IEEEbib}
\bibliography{strings,refs}

\end{document}